\documentclass[runningheads]{llncs}
\usepackage{graphicx}
\usepackage{cite}
\usepackage{amsmath,amssymb,amsfonts}
\usepackage{algorithmic}
\usepackage{textcomp}
\usepackage{xcolor}
\usepackage{verbatim}
\usepackage{booktabs}
\usepackage{multirow}
\usepackage{subcaption}
\usepackage{soul}
\usepackage{floatrow}
\begin{document}
\title{Classifying Scientific Publications with BERT - Is Self-Attention a Feature Selection Method?%\thanks{Supported by organization x.}
}
\titlerunning{Is Self-Attention a Feature Selection Method?}
\author{Andres Garcia-Silva\inst{1}\orcidID{0000-0002-5664-488X} \and
Jose Manuel Gomez-Perez\inst{1}\orcidID{0000-0002-5491-6431} 
}
\authorrunning{A. Garcia-Silva and J. M. Gomez-Perez}
\institute{expert.ai Research Lab \and
Prof. Waskman 10, 28036 Madrid, Spain \\
\email{\{agarcia,jmgomez\}@expert.ai}\\
\url{https://www.expert.ai} }

\maketitle              % typeset the header of the contribution
\begin{abstract}
We investigate the self-attention mechanism of BERT in a fine-tuning scenario for the classification of scientific articles over a taxonomy of research disciplines. We observe how self-attention focuses on words that are highly related to the domain of the article. Particularly, a small subset of vocabulary words tends to receive most of the attention. We compare and evaluate the subset of the most attended words with feature selection methods normally used for text classification in order to characterize self-attention as a possible feature selection approach. Using ConceptNet as ground truth, we also find that attended words are more related to the research fields of the articles. However, conventional feature selection methods are still a better option to learn classifiers from scratch. This result suggests that, while self-attention identifies domain-relevant terms, the discriminatory information in BERT is encoded in the contextualized outputs and the classification layer. It also raises the question whether injecting feature selection methods in the self-attention mechanism could further optimize single sequence classification using transformers.

%Se puede considerar el attention de los transformers como un feature selection?
%Se puede evaluar los metodos de feature selection?
%Observamos que las words son mos relacionadas con las dominios
%Qualitative analisis: Domain Specific
%Medir la estabilidad. 
%Evaluar otros metodos de clasificación.
%Kalousis A., Prados J., Hilario M.
%Stability of feature selection algorithms: A study on high-dimensional spaces
%Knowledge & Information Systems, 12 (1) (2007), 

\keywords{Neural Language Models \and Text Classification \and Scholarly Communications.}
\end{abstract}

\section{Introduction}

The annotation and classification of scientific literature is a crucial task to make scientific knowledge easily discoverable, accessible, and reusable, accelerating scientific breakthroughs by helping scholars locate and understand the right research, making connections, and overcoming information overload. Some examples of efforts to structure scientific literature include scientific search engines like Semantic Scholar \cite{Ammar2018ConstructionLiteratureGraph} and Microsoft Academic \cite{ArnabMicrosoftAcademicGraph2015}. Both rely on knowledge graphs to enable a structured representation of scientific knowledge that supports applications like topic-driven search and recommendation. Similarly, scientific publishers have released knowledge graphs such as SN SciGraph \cite{HammondPT17:SciGraphModeling} in order to more effectively organize their publications and increase automation. Other efforts like ORKG\cite{ORKG} rely on knowledge graphs to structure the actual contributions described in the publications, making research results on a specific topic comparable across the literature. 

Publications are therefore being annotated with information about their content, which includes topics \cite{Ammar2018ConstructionLiteratureGraph}, fields of study \cite{ArnabMicrosoftAcademicGraph2015}, concepts \cite{HammondPT17:SciGraphModeling}, and research fields \cite{ORKG}. Such metadata is generally based on controlled vocabularies and arranged according to a taxonomy \cite{HammondPT17:SciGraphModeling, ORKG}, thesaurus \cite{Ammar2018ConstructionLiteratureGraph, ArnabMicrosoftAcademicGraph2015} or ontology \cite{topicminer2}. In some cases, the annotation process can be fully automatic \cite{Ammar2018ConstructionLiteratureGraph, ArnabMicrosoftAcademicGraph2015}. However, authors are often asked to manually classify their contribution in the right categories, which is tedious and error-prone. In other occasions, this task falls under the responsibility of a reduced number of senior expert editors, making the process expensive and slow \cite{topicminer2}. 

In this paper, we focus on the task of classifying scientific publications against a taxonomy of scientific disciplines. A wide variety of approaches are suitable for this task, including machine learning classifiers that rely on high-dimensional sparse representations \cite{Joachims2006LinearSVM}, deep learning classifiers using dense representations \cite{joulin2016FastTextClass}, and rule-based or heuristic methods \cite{topicminer2}. Encouraged by the success of recent developments in natural language processing and understanding, where pre-trained transformer language models dominate the state of the art~\cite{wang2019glue}, herein we focus on BERT \cite{devlin2018BERT} and its different flavors specialized in the scientific domain: BioBERT \cite{lee2020biobert} and SciBERT \cite{Beltagy2019SciBERT}. 

Our experiments confirm that using transformers to train scientific classifiers generally results in greater accuracies compared to linear classifiers that were until now regarded as strong baselines \cite{joulin2016FastTextClass}. We also observe that fine-tuning pre-trained transformers on domain-specific corpora contributes to this goal. However, despite previous research focused on interpreting and understanding how transformers encode information \cite{tenney-etal-2019-bert, jawahar-etal-2019-bert, ClarkWhatDoesBERTLookAt2019, kovaleva2019revealing, Rogers2020Bertology}, the actual mechanism by which fine-tuning impacts on our classification task is still unclear. In an effort to shed light on this matter, we focus on analyzing the self-attention mechanism inherent of the transformer architecture \cite{Vaswani2017Transformers}. Our findings show that the last layer of BERT attends to words that are semantically relevant for the scientific fields associated with each publication. This observation suggests that self-attention actually performs some type of feature selection for the fine-tuned model. 

We investigate the possible relation between self-attention and feature selection methods from different perspectives, including vocabulary overlap, ranking similarity, domain relevance, feature stability, and classification performance. Our results open a future research path to determine whether injecting feature selection methods in the self-attention mechanism could derive even better results for single sequence classification using transformer architectures. 

Our main contributions in this paper are the following:
\begin{itemize}
\item We leverage the vertical pattern present in the transformer self-attention mechanism of BERT, SciBERT and BioBERT, where some words receive more attention on average than the rest of the words, and compare it against conventional feature selection methods used in text classification.
\item We find that self-attention has interesting properties as a feature selection method. The most attended words are in general more relevant to the publication domain than those found using conventional approaches to feature selection. The stability of the features resulting from self-attention is in line with the results obtained through conventional approaches. However, when used to learn classifiers from scratch, methods like chi-square and information gain contribute to train better classifiers.  
\item We analyze from a semantic point of view the self-attention mechanism and quantify the amount of domain knowledge it encodes in the hidden states of the last layer. To this purpose, we rely on ConceptNet \cite{speer2017conceptnet}, a commonsense knowledge graph where attended words are mapped to concepts from which we derive their corresponding domains. 
\end{itemize}

The remainder of the paper is structured as follows. Section 2 describes related work in the annotation of scientific publications, classification, transformer language models, and other work focused on the analysis of transformer self-attention. In section 3, we present  experimental results classifying research papers into a scientific taxonomy. In section 4, we motivate the analysis of self-attention as feature selection with examples of attended words and scientific categories. In section 5, we quantify the relation between self-attention and feature selection methods. Finally, section 6 concludes the paper\footnote{Tables, datasets and notebooks to reproduce our experiments are available in \url{https://github.com/expertailab/Is-BERT-self-attention-a-feature-selection-method}}.

\section{Related Work}
\label{sec:rel}
Annotating research articles with entities such as research fields or topics is addressed in the literature using entity recognition and similarity measures between entity labels and their mentions \cite{ChernyakAnnotationResearch}. In Microsoft Academic Graph \cite{ArnabMicrosoftAcademicGraph2015} the candidate entities (field of study) are identified using string matching between the entity keywords and their paper mentions, then rules are applied to gather more candidates and to filter out the less relevant entities. Similarly, the CSO classifier \cite{topicminer2}, which assigns articles to concepts in the Computer Science Ontology\footnote{See \url{http://cso.kmi.open.ac.uk/}}, first identifies concepts explicitly mentioned in the text and then, in an effort to find entities not explicitly mentioned, it uses a similarity measure based on word embeddings. 
In the Semantic Scholar literature graph \cite{Ammar2018ConstructionLiteratureGraph}, an ensemble of tools is used to annotate entities: statistical models for entity span prediction and disambiguation, rules for string-based entity spotting, and off-the-shelf tools\footnote{\url{https://sobigdata.d4science.org/web/tagme/tagme-help}}. 

In addition, different models can be used for this task, including SVM\cite{Joachims2006LinearSVM} or softmax classifiers \cite{featseltextclass2020}. Mai et al. \cite{MaiDeepLearningSubject} proposed classifiers based on convolutional\cite{Kim2014ConvolutionalNN} and recurrent neural networks \cite{yin2017comparativeCNNandRNN} to annotate research articles. However, such deep learning classifiers need to be trained from scratch and depend on the network architecture. On the contrary, neural language models and particularly transformers like GPT-2\cite{radford2019language} or BERT \cite{devlin2018BERT} are pre-trained on a large corpus and then fine-tuned for classification by just adding a linear classifier to the model output. This approach has proven to successfully tackle several NLP tasks \cite{wang2019glue}, including text classification. In the scientific domain, SciBERT\cite{Beltagy2019SciBERT} and BioBERT\cite{lee2020biobert} have also reported state of the art results. Researchers are investigating the mechanics underlying BERT~\cite{Rogers2020Bertology}, analyzing its hidden states and outputs \cite{tenney-etal-2019-bert, jawahar-etal-2019-bert}, as well as the self-attention mechanism \cite{kovaleva2019revealing, ClarkWhatDoesBERTLookAt2019}. Unlike previous approaches \cite{kovaleva2019revealing, ClarkWhatDoesBERTLookAt2019}, we semantically analyze the words that are attended above average in the last hidden state, leveraging the commonsense knowledge represented in ConceptNet, and quantify the relation between attention and feature selection methods often used in text classification. 

\section{Fine-tuning Language Models for Text Classification}
We evaluate the use of language models on a text classification task where research articles are labeled with one or more knowledge fields. To this purpose, we choose: i) BERT and GPT-2, pre-trained on a  general-purpose corpus, ii) SciBERT, pre-trained solely on scientific documents, and iii) BioBERT, pre-trained on a combination of general and scientific text. Table \ref{tab:lmsdetails}, provides relevant information about each language model, its pre-training and vocabulary. BioBERT uses the same tokenization method and vocabulary as BERT, while SciBERT adopts SentencePiece, based on WordPiece tokenization. The overlap between the vocabularies of BERT and SciBERT is 42\%, which shows a substantial difference in the most frequently used words in the scientific domain and general-purpose documents. We choose the base version of BERT models (12 layers, 768 hidden size, 12 attention heads per layer) and a comparable model for GPT-2. 

\begin{table}[htbp]
  \centering
  \caption{Language models pre-training information.}
   \resizebox{\textwidth}{!}{
% Table generated by Excel2LaTeX from sheet 'Pretrained LM'
        \begin{tabular}{lp{6.665em}p{3.5em}p{17.28em}cc}
        \toprule
        Model & \multicolumn{1}{c}{Tokenizer} & \multicolumn{1}{c}{Vocabulary} & \multicolumn{1}{c}{Corpus} & Domains & steps/epochs \\
        \midrule
        BERT  & \multicolumn{1}{c}{WordPIece} & \multicolumn{1}{c}{30K}   & BookCorpus (2.5B tokens) + Wikipedia (0.8B tokens) & General & 1M steps \\
        BioBERT 1.1 & \multicolumn{1}{c}{WordPiece } & \multicolumn{1}{c}{BERT} & BERT corpus  + PubMed abstracts (4.5B tokens) & General + Biomedic & 1M steps \\
        BioBERT 1.0 & \multicolumn{1}{c}{WordPiece} & \multicolumn{1}{c}{BERT} & BERT Corpus + PubMed abstracts (4.5B tokens) + PMC full-text articles (13.5M tokens)  & General + Biomedic & 470K steps \\
        SciBERT & \multicolumn{1}{c}{SentencePiece } & \multicolumn{1}{c}{30K}   & Semantic Scholar (3.17B tokens) (1.14M  full text papers) & \multicolumn{1}{p{9.945em}}{18\% Computer Science and 82\% Biomedical} & Not reported \\
        GPT-2  & Byte Pair Encoding (BPE) & \multicolumn{1}{c}{50k}   & 8 million web pages, except Wikipedia (40GB of text) & General & Not reported \\
        \bottomrule
        \end{tabular}%
    }
  \label{tab:lmsdetails}%
\end{table}%

To fine-tune BERT, BioBERT and SciBERT on our multilabel classification task, we follow the guidelines provided by Devlin. et al. \cite{devlin2018BERT} for single-sentence classification. We take the last layer encoding of the classification token $<$CLS$>$ and add an N-dimensional linear layer, with N the number of classification labels. We use a binary cross-entropy loss function  
to allow the model to assign independent probabilities to each label. For GPT-2 we also add a linear layer on top of the last hidden state for the classification token. We train the models for 4 epochs, with batch size 8 and 2e-5 learning rate. 

As a baseline, we use an SVM with a linear kernel \cite{Fan2008LibLinear}. We follow a one-vs-all strategy to train a binary SVM classifier per category, with grid search for the regularization parameter. We use WordNet to lemmatize the words, whenever they exist in the WordNet lexicon, and remove stop words. In addition, we use fastText \cite{joulin2016FastTextClass} to learn a hierarchical softmax classifier using n-gram embeddings. We learn binary classifiers for each category, with automatic hyperparameter optimization to fix learning rate, number of epochs, and n-gram length. 

We gather our dataset of scientific articles from a broad range of knowledge fields in SciGraph~\cite{HammondPT17:SciGraphModeling},  
where articles are labelled following the ANZSRC\footnote{Australian and New Zealand Standard Research Classification.} taxonomy. This taxonomy comprises 22 first level categories, such as \textit{Economics}, \textit{Law}, and \textit{Computer Science}, each of them with their own subcategory tree.  From SciGraph, we extract the titles and abstracts of articles published in 2011 and 2012, as well as their categories. In total, we gather 405K papers, 187K from 2011 and the rest from 2012. In average, each first level category has 20,164 articles with a standard deviation of 31,791, which shows how unevenly the different categories are covered. Some of them are well represented, like \textit{Medical And Health Sciences}, with 138,728 articles, while others, like \textit{Studies In Creative Arts And Writing}, have little over a hundred articles. 

%\subsection{Experimental results}
We fine-tune the language models to learn to classify papers on any of the 22 first level categories. We train on papers only from 2011 and evaluate using  5-fold cross validation. Table \ref{tab:res1} shows that the transformers pre-trained on a  scientific corpus generally achieve greater f-measure in this task. The exception is BioBERT-1.0, which scores under BERT. BioBERT-1.0 was pre-trained on a lower number of steps than the other transformers, which could be affecting its performance. GPT-2 is the model producing the lowest f-measure, which shows evidence of a potential mismatch between the vocabulary and quality of the  scientific corpus and the Web corpus where it was pre-trained, which may be undermining its performance. Overall, transformers produce more accurate classifiers than the linear methods used as baselines.

\begin{table}
    \centering
    \caption{Evaluation results of the multilabel classifiers (f-measure) on first level categories (a), and on second level categories (b).}
    \resizebox{\textwidth}{!}{
% Table generated by Excel2LaTeX from sheet 'ClassifierResults'
\begin{tabular}{rrp{1em}rrrccccccc}
\multicolumn{2}{c}{First level categories} &       & \multicolumn{8}{c}{Second level categories}                   &  \\
\cmidrule{1-2}\cmidrule{5-12}\multicolumn{1}{c}{\textbf{Model}} & \multicolumn{1}{c}{\textbf{f-measure}} &       &       & \textbf{Categories} & \textbf{Articles} & \textbf{subcat.} & \textbf{Bert} & \textbf{BioBERT-1.1} & \textbf{SciBERT} & \textbf{SVM} & \textbf{fastText} \\
\cmidrule{1-2}\cmidrule{5-12}\multicolumn{1}{c}{SciBERT} & \multicolumn{1}{c}{\textbf{0.838}} &       &       & Biological  & 65340 & 9     & 0.883 & 0.884 & \textbf{0.887} & 0.880 & 0.871 \\
\multicolumn{1}{c}{BioBERT-1.1} & \multicolumn{1}{c}{0.825} &       &       & Medical and Health & 58068 & 18    & 0.838 & 0.843 & \textbf{0.854} & 0.836 & 0.819 \\
\multicolumn{1}{c}{BERT} & \multicolumn{1}{c}{0.819} &       &       & Chemical  & 40837 & 8     & 0.858 & 0.862 & \textbf{0.865} & 0.854 & 0.847 \\
\multicolumn{1}{c}{BioBERT-1.0} & \multicolumn{1}{c}{0.818} &       &       & Mathematical & 28723 & 5     & 0.886 & 0.883 & \textbf{0.891} & 0.884 & 0.878 \\
\multicolumn{1}{c}{GPT-2} & \multicolumn{1}{c}{0.808} &       &       & Computer Sciences & 20777 & 6     & 0.861 & 0.862 & \textbf{0.864} & 0.861 & 0.849 \\
\multicolumn{1}{c}{SVM} & \multicolumn{1}{c}{0.807} &       &       & Language & 2233  & 6     & \textbf{0.911} & 0.900 & 0.903 & 0.900 & 0.906 \\
\multicolumn{1}{c}{fastText} & \multicolumn{1}{c}{0.790} &       &       & Hist. And Archelogy & 2076  & 4     & \textbf{0.955} & 0.950 & 0.941 & 0.946 & 0.946 \\
      &       &       &       & Built Environment & 140   & 4     & 0.495 & 0.700 & 0.697 & \textbf{0.808} & 0.804 \\
      &       &       &       & Creative Arts & 132   & 4     & 0.639 & 0.788 & 0.781 & \textbf{0.925} & 0.828 \\
\cmidrule{1-2}\cmidrule{5-12}\end{tabular}%

    }
    \label{tab:res1}
\end{table}

To further explore the relation between the pre-training and fine-tuning corpora, we learn classifiers to label articles with second level categories in ANZSRC for some of the first level categories. For this experiment, we enlarge our dataset with articles published in 2012 and evaluate only the best language models, discarding BioBERT 1.0 and GPT-2. The results in table \ref{tab:res1} show that, in general, scientific categories are dominated by SciBERT and BioBERT-1.1. However, for categories in humanities, e.g \textit{Language}, and \textit{History and Archaeology}, BERT produces better classifiers, providing evidence that the general-purpose knowledge encoded in BERT is more relevant in those cases. Interestingly, when there are few examples, e.g., in categories \textit{Built Environment} and \textit{Creative Arts}, the general knowledge encoded in BERT is of little use for the classifiers, while the scientific knowledge in BioBERT-1.1 and SciBERT contributes to achieve higher f-measure. Linear classifiers outperform transformer-based models in such under-represented categories. 

\section{Exploring self-attention heads}
Above we show that BERT-based models are able to produce high performance multilabel classifiers. However, we know little about what makes them good at this task. In this section, we inspect the self-attention mechanism underpinning such models as a key element to understanding this behavior. 

\begin{figure*}[ht!]
\footnotesize
    \begin{subfigure}[t]{.49\linewidth}
        \centering
        \includegraphics[width=\linewidth]{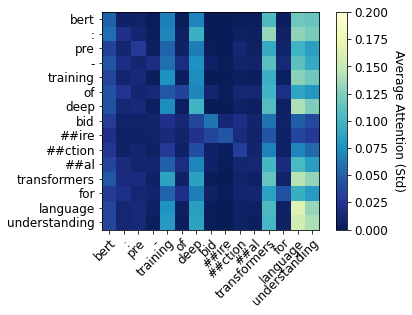}
        \caption{BERT}
        \label{fig:attBert1}
    \end{subfigure}
    \begin{subfigure}[t]{.49\linewidth}
        \centering
        \includegraphics[width=\linewidth]{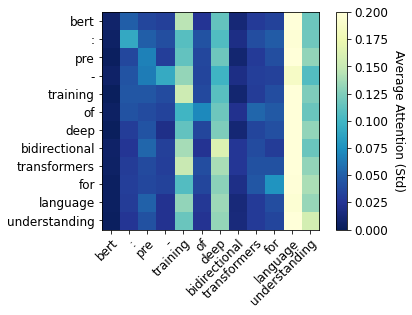}
        \caption{SciBERT}
        \label{fig:attSciBERT1}
    \end{subfigure}%
    \\
    \begin{subfigure}[t]{.48\linewidth}
        \centering
        \includegraphics[width=\linewidth]{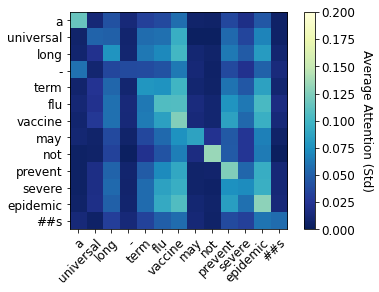}
        \caption{BioBERT-1.1}
        \label{fig:attBioBERT2}
    \end{subfigure}
    \begin{subfigure}[t]{.48\linewidth}
        \centering
        \includegraphics[width=\linewidth]{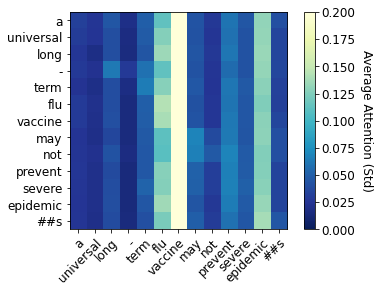}
        \caption{SciBERT}
        \label{fig:attSciBERT2}
    \end{subfigure}%    
    \caption{Average weights in the self-attention heads of the last hidden state.}
    \label{fig:attentionavg}
\end{figure*}

According to Clark et al. \cite{ClarkWhatDoesBERTLookAt2019}, attention weights indicate how relevant a particular word is when computing the next representation for the current word. To illustrate this statement, figure \ref{fig:attentionavg}, depicts the mean weights of the 12 self-attention heads in the last hidden state of the fine-tuned models for two papers titled \textit{"BERT: Pre-training of Deep Bidirectional Transformers for Language Understanding"}, and \textit{"A universal long-term flu vaccine may not prevent severe epidemics"}. The plots clearly show the so-called vertical pattern \cite{kovaleva2019revealing}, where a few tokens receive most of the attention, such as \textit{training, deep, transformer, language}, and \textit{understanding} in the first sentence, and \textit{flu, vaccine, prevent, severe} and \textit{epidemic} in the second. Note how while the vocabulary captured by SciBERT includes the word \textit{bidirectional}, BERT uses subwords to represent it.

We do not include special tokens $<$SEP$>$ and $<$CLS$>$ since the amount of attention received by these tokens makes the attention received by the other tokens barely noticeable.  Clark et al.~\cite{ClarkWhatDoesBERTLookAt2019} speculate that the attention on $<$SEP$>$ in one head could indicate that the attention heads function is not applicable, while Rogers et al.~\cite{Rogers2020Bertology} interpret the attention on $<$CLS$>$ as the attention on a pooled sentence-level representation. 

From these two examples, we observe that the most attended words in the last hidden state are highly related to the research fields of the articles: \textit{Computer science} and \textit{Medical and Health Sciences}. So, we look into this relation and identify the words that receive most vertical attention in the last hidden state for a subset of our dataset where each first level category is represented with at most 500 papers. First, for each input sequence we calculate the mean weights for the 12 attention heads in the last hidden state. Next, we generate a new weight matrix grouping subwords into words by averaging the subword weights. Finally, we gather the words with a vertical mean attention above the mean attention in the weight matrix. This results in 8,840 attended words for BERT, 17,773 for BioBERT, and 12,265 for SciBERT, corresponding to 16\%, 32\%, and 22\% of the vocabulary managed by each language model.

Table \ref{tab:mostattendedwords} shows the top 20 most frequent attended words in three research fields: \textit{Biology}, \textit{Computer Science} and \textit{History and Archaeology}. As can be noted, most of such words are highly related to the specific research field, appearing along a few punctuation marks and some stop words. While frequent attention to periods and commas was already reported in \cite{kovaleva2019revealing, ClarkWhatDoesBERTLookAt2019}, the reason why this happens is not clear yet. Rogers et al. \cite{Rogers2020Bertology} suggest that it must be related to model overparameterization while Clark et al. \cite{ClarkWhatDoesBERTLookAt2019} point at the high frequency of these tokens in the corpus. Stop words are also highly frequent words and the models could be learning to attend to them as in the case of punctuation marks. 

\begin{table}[htbp]
  \centering
  \caption{Most attended words above average attention in the fine-tuned models.}
  \resizebox{\textwidth}{!}{%
    \begin{tabular}{lllrlllrlll}
    \toprule
    \multicolumn{3}{c}{06 - Biological Sciences } &       & \multicolumn{3}{c}{08 - Computer Science} &       & \multicolumn{3}{c}{21 - History and Archaeology} \\
\cmidrule{1-3}\cmidrule{5-7}\cmidrule{9-11}    \multicolumn{1}{c}{BERT} & \multicolumn{1}{c}{BioBERT} & \multicolumn{1}{c}{SciBERT} &       & \multicolumn{1}{c}{BERT} & \multicolumn{1}{c}{BioBERT} & \multicolumn{1}{c}{SciBERT} &       & \multicolumn{1}{c}{BERT} & \multicolumn{1}{c}{BioBERT} & \multicolumn{1}{c}{SciBERT} \\
\cmidrule{1-3}\cmidrule{5-7}\cmidrule{9-11}     ,    &  of   &  .    &       &  ,    &  the  &  .    &       &  ,    &  the  &  .  \\
     species  &  the  &  ,    &       &  .    &  of   &  ,    &       &  .    &  of   &  ,  \\
     gene  &  in   &  gene  &       &  data  &  data  &  data  &       &  history  &  in   &  history  \\
     cell  &  .    &  species  &       &  )    &  -    &  information  &       &  )    &  history  &  century  \\
     cells  &  to   &  cell  &       &  image  &  time  &  algorithm  &       &  historical  &  -    &  historical  \\
     protein  &  species  &  the  &       &  network  &  information  &  network  &       &  archaeological  &  to   &  modern  \\
     .    &  and  &  protein  &       &  information  &  model  &  image  &       &  cultural  &  century  &  the  \\
     genetic  &  for  &  expression  &       &  networks  &  system  &  algorithms  &       &  the  &  .    &  archaeological  \\
     plants  &  -    &  genes  &       &  control  &  algorithm  &  as   &       &  social  &  and  &  social  \\
     plant  &  gene  &  genetic  &       &  images  &  in   &  networks  &       &  archaeology  &  a    &  cultural  \\
     expression  &  a    &  cells  &       &  algorithms  &  systems  &  systems  &       &  political  &  historical  &  american  \\
     growth  &  cell  &  growth  &       &  software  &  based  &  model  &       &  culture  &  period  &  human  \\
     genes  &  protein  &  plants  &       &  neural  &  a    &  analysis  &       &  women  &  early  &  literature  \\
     )    &  genes  &  plant  &       &  optimization  &  network  &  software  &       &  literary  &  modern  &  data  \\
     molecular  &  cells  &  dna  &       &  simulation  &  to   &  time  &       &  heritage  &  archaeological  &  state  \\
     dna  &  on   &  proteins  &       &  learning  &  algorithms  &  images  &       &  precipitation  &  world  &  women  \\
     stress  &  with  &  molecular  &       &  search  &  analysis  &  control  &       &  education  &  social  &  life  \\
     populations  &  expression  &  populations  &       &  web  &  image  &  simulation  &       &  identity  &  on   &  period  \\
     population  &  genetic  &  population  &       &  a    &  models  &  problems  &       &  literature  &  years  &  political  \\
     genome  &  plants  &  water  &       &  classification  &  user  &  such  &       &  past  &  american  &  development  \\
     \bottomrule
    \end{tabular}%
    }
  \label{tab:mostattendedwords}%
\end{table}%

\section{Feature Selection}
In the previous section we show that fine-tuned BERT models concentrate their attention on a subset of the overall vocabulary that ranges between 16\% to 32\% of the words. Following this observation, we hypothesize that such attention on a selected fragment of the vocabulary is the transformer version of feature selection. However, rather than picking the most interesting features for a classifier, self-attention selects words that heavily influence the representation of the rest of the words in the same sequence. We investigate whether there is a relation between feature selection algorithms commonly used for text classification and the most attended words in the fine-tuned language models. 

We center our analysis on four feature selection methods used for text classification \cite{featseltextclass2020,introInfoRetrieval2008, Simeon2008CategoricalPD}: Chi-square  (chi), Information Gain (ig), Document Frequency (df), and Categorical Proportional Difference (pd). Chi-square measures the lack of independence between a word and a class; its value is zero if the word and the class are independent. Information Gain measures the entropy reduction of the dataset when it is split by a feature value. Thus, words with larger information gain discriminate the data ensuring a lower entropy. Document Frequency counts the number of documents where a term appears. Categorical Proportional Difference measures the degree to which a word contributes to differentiating a particular category from others. %If the word only occurs in one class the value is 1.0.
% Mutual information measures the amount of information the presence or absence of a word contributes to each class. MI reaches its maximum value if the word is a perfect indicator for a class.   %If the word only occurs in one class the value is 1.0. \cite{Yang1997FeatureSelectioninText} 

\begin{figure}[htb]
\CenterFloatBoxes
\begin{floatrow}
\ttabbox[.3\textwidth]
  { \resizebox{0.18\textwidth}{!}{
% Table generated by Excel2LaTeX from sheet 'Overlap'
\begin{tabular}{clc}
\toprule
\textbf{LM} & \multicolumn{1}{c}{\textbf{FS}} & \textbf{\%} \\
\midrule
\multirow{4}[2]{*}{BERT} &  dc   & 60\% \\
      &  ig   & 54\% \\
      &  chi   & 43\% \\
      &  pd   & 12\% \\      
\cmidrule{2-3}\multirow{4}[2]{*}{BioBERT-1.1} &  dc   & 64\% \\
      &  ig   & 55\% \\
      &  chi   & 44\% \\
      &  pd   & 25\% \\      
\cmidrule{2-3}\multirow{4}[2]{*}{SciBERT} &  dc   & 58\% \\
      &  ig   & 49\% \\
      &  chi   & 42\% \\
      &  pd   & 20\% \\      
\bottomrule
\end{tabular}%

    }
  }
  {\caption{Word overlap: most attended vs. feature selection.}\label{tab:fs1}}
\killfloatstyle
\ffigbox[.7\textwidth]
  {\includegraphics[width=.7\textwidth]{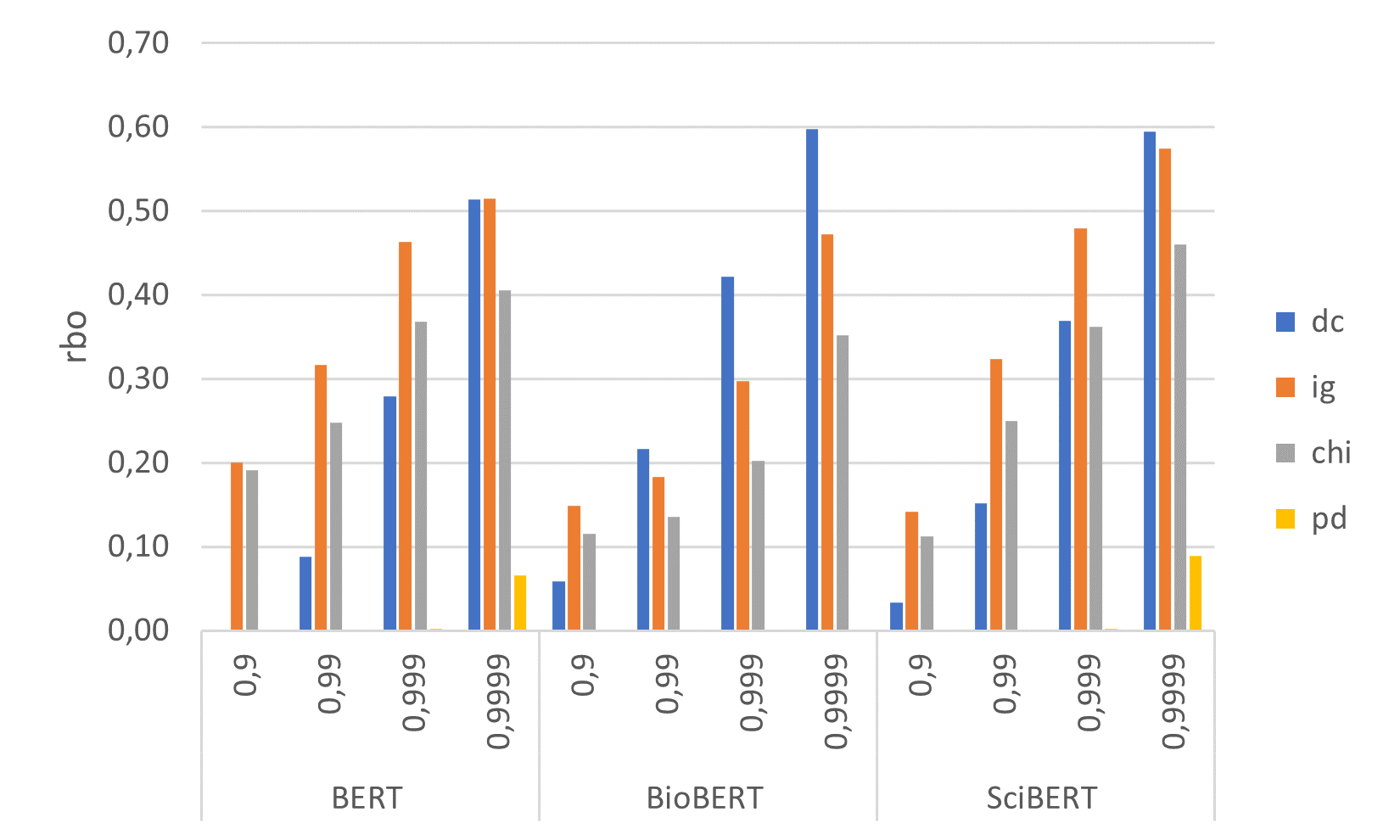}}
  {\caption{Rank-biased overlap at different p values between most attended words and feature selection algorithms.}\label{fig:fs1rbo}}
\end{floatrow}
\end{figure}

We compare the most attended words with those selected by the above-mentioned feature selection methods, and measure how similar the rankings of words sorted by their average attention are to the rankings produced by each feature selection method. In table \Ref{tab:fs1}, we report the vocabulary overlap of the most attended words and feature selection methods after filtering out the stop words. The number of features selected was limited to the top k words, where k is the number of words attended above average by each language model. Indeed, the results indicate a large overlap. Fine-tuned language models for text classification attend up to 64\% of the common terms returned by dc, the most simple of our feature selection baselines, which itself performs similarly to ig and chi \cite{Yang1997FeatureSelectioninText}. For all three models, their most attended words have the largest overlap with document frequency, followed by information gain, chi-square and, finally, proportional difference.

To measure the similarity between rankings we apply the Rank-Biased Overlap  (RBO) \cite{webber2010RankBiasedOverlap} metric. RBO ranges between 0 to 1, from less to more similar, and was designed for non-conjoint rankings, i.e. both lists may have different items, may be incomplete and with different length. Through the \textit{p} parameter, RBO models the probability to continue considering the overlap at the next rank, having examined the overlap at the previous rank. Figure \ref{fig:fs1rbo} shows the RBO for the attention and feature selection rankings. We set p to 0.9, 0.99, 0.999, and 0.9999, indicating the model to assign the first 10, 100, 1,000, and 10,000 ranks respectively, approximately 85\% to 86\% of the weight of the evaluation. 

While the BERT and SciBERT attended words rankings are more similar to the ranking of discriminative words (ig) for p values of 0.9 to 0.999, they finally converge with the ranking of common terms (dc), too. On the other hand, the BioBERT-1.1 ranking is clearly most similar to the common term rankings (dc). We think that the difference between the three models could be related to the subword vocabulary and pre-training corpus. Subword vocabularies are tightly related to the training corpus since they are generated to represent the whole corpus with the minimum number of word pieces. BERT trains its own subword vocabulary on a general corpus and during fine-tuning learns to attend more to discriminative words in the scientific domain. SciBERT also uses its own vocabulary trained on a limited scientific corpus, enabling the model to attend to discriminative words (like BERT) but also to common words due to the domain knowledge it encodes. BioBERT on the other hand reuses the BERT subword vocabulary and therefore many scientific terms are split in a suboptimal number of pieces. This has a negative impact on the ability of the self-attention mechanism to focus on discriminative words, and subsequently on the attention to common terms. 

\subsection{Domain knowledge}
We investigate the domain relevance of the words that are most attended by the language models and compare it with words produced by the feature selection methods. To this end, we search the words in ConceptNet and leverage the relation \textit{HasContext} to identify the domains where they are commonly used. We manually map the 22 first level categories in ANZSRC to the corresponding concepts in ConceptNet. To deal with morphological variations like plurals and conjugations we use the \textit{FormOf} relation, and to increase the coverage we traverse the \textit{isA} type hierarchy one level up looking for the corresponding concept. For example, the word \textit{networking} is a \textit{FormOf} of the root word \textit{network}, which in turn \textit{HasContext} \textit{Computer Science} and \textit{Electronics}, and the concept \textit{Electronics} \textit{isA} type of \textit{Physics}. 

% Table generated by Excel2LaTeX from sheet 'ConceptNet'
\begin{table}[htbp]
  \centering
  \caption{words per category matching the corresponding ConceptNet context.}
  \footnotesize
  \resizebox{\textwidth}{!}{%
% Table generated by Excel2LaTeX from sheet 'ConceptNET Lemmas (final)'
\begin{tabular}{rrcccccccccccccccccccccc}
\toprule
      &       & \multicolumn{3}{c}{\textbf{Mean Self-Att.}} &       & \multicolumn{4}{c}{\textbf{Feat. Sel.}} &       & \multicolumn{3}{c}{\textbf{Self-Att. (TF)}} &       & \multicolumn{4}{c}{\textbf{TF}} &       & \multicolumn{4}{c}{\textbf{TF/IDF}} \\
\cmidrule{3-5}\cmidrule{7-10}\cmidrule{12-14}\cmidrule{16-19}\cmidrule{21-24}\multicolumn{1}{c}{\textbf{Category}} &       & \textbf{BERT} & \textbf{BioB.} & \textbf{SciB.} &       & \textbf{dc} & \textbf{ig} & \textbf{chi} & \textbf{pd} &       & \textbf{BERT} & \textbf{BioB.} & \textbf{SciB.} &       & \textbf{dc} & \textbf{ig} & \textbf{chi} & \textbf{pd} &       & \textbf{dc} & \textbf{ig} & \textbf{chi} & \textbf{pd} \\
\cmidrule{1-1}\cmidrule{3-5}\cmidrule{7-10}\cmidrule{12-14}\cmidrule{16-19}\cmidrule{21-24}\multicolumn{1}{c}{Mathematics} &       & 36    & 18    & 28    &       & 29    & 18    & 16    & \multicolumn{1}{r}{25} &       & 60    & 54    & 53    &       & 51    & 52    & 53    & \multicolumn{1}{r}{33} &       & 53    & 53    & 55    & \multicolumn{1}{r}{35} \\
\multicolumn{1}{c}{Physics} &       & 21    & 4     & 22    &       & 18    & 11    & 13    & \multicolumn{1}{r}{20} &       & 41    & 42    & 38    &       & 33    & 33    & 38    & \multicolumn{1}{r}{18} &       & 41    & 41    & 42    & \multicolumn{1}{r}{18} \\
\multicolumn{1}{c}{Chemistry} &       & 20    & 7     & 18    &       & 7     & 15    & 16    & \multicolumn{1}{r}{20} &       & 29    & 30    & 27    &       & 24    & 24    & 25    & \multicolumn{1}{r}{36} &       & 27    & 27    & 29    & \multicolumn{1}{r}{37} \\
\multicolumn{1}{c}{Biology} &       & 18    & 6     & 18    &       & 11    & 15    & 14    & \multicolumn{1}{r}{11} &       & 44    & 43    & 38    &       & 25    & 24    & 28    & \multicolumn{1}{r}{14} &       & 34    & 33    & 35    & \multicolumn{1}{r}{16} \\
\multicolumn{1}{c}{Agriculture} &       & 1     & 1     & 0     &       & 0     & 0     & 0     & 1     &       & 4     & 4     & 4     &       & 1     & 1     & 1     & 0     &       & 3     & 3     & 3     & \multicolumn{1}{r}{0} \\
\multicolumn{1}{c}{Comp. Science} &       & 6     & 4     & 7     &       & 11    & 5     & 5     & 4     &       & 20    & 17    & 18    &       & 14    & 14    & 16    & 11    &       & 15    & 15    & 16    & \multicolumn{1}{r}{12} \\
\multicolumn{1}{c}{Technology} &       & 5     & 0     & 3     &       & 1     & 1     & 1     & 0     &       & 3     & 2     & 2     &       & 1     & 1     & 1     & 1     &       & 1     & 1     & 2     & \multicolumn{1}{r}{0} \\
\multicolumn{1}{c}{Medicine} &       & 16    & 13    & 22    &       & 11    & 12    & 15    & 11    &       & 30    & 28    & 32    &       & 19    & 19    & 20    & 17    &       & 21    & 21    & 22    & \multicolumn{1}{r}{20} \\
\multicolumn{1}{c}{Education} &       & 2     & 1     & 1     &       & 1     & 1     & 1     & 3     &       & 4     & 8     & 6     &       & 4     & 4     & 4     & 1     &       & 5     & 5     & 5     & \multicolumn{1}{r}{4} \\
\multicolumn{1}{c}{Economics} &       & 2     & 4     & 1     &       & 1     & 2     & 2     & 0     &       & 8     & 10    & 9     &       & 8     & 8     & 7     & 0     &       & 9     & 9     & 9     & \multicolumn{1}{r}{0} \\
\multicolumn{1}{c}{Commerce} &       & 7     & 4     & 2     &       & 0     & 1     & 1     & 0     &       & 6     & 6     & 7     &       & 3     & 3     & 4     & 2     &       & 6     & 6     & 6     & \multicolumn{1}{r}{2} \\
\multicolumn{1}{c}{Psychology} &       & 2     & 2     & 0     &       & 4     & 1     & 0     & 2     &       & 8     & 7     & 5     &       & 6     & 6     & 7     & 7     &       & 9     & 9     & 9     & \multicolumn{1}{r}{7} \\
\multicolumn{1}{c}{Law} &       & 5     & 6     & 2     &       & 6     & 3     & 4     & 7     &       & 9     & 7     & 8     &       & 9     & 9     & 8     & 7     &       & 8     & 8     & 9     & \multicolumn{1}{r}{8} \\
\multicolumn{1}{c}{Literature} &       & 1     & 0     & 0     &       & 1     & 1     & 0     & 2     &       & 0     & 0     & 1     &       & 1     & 1     & 1     & 1     &       & 0     & 0     & 0     & \multicolumn{1}{r}{1} \\
\multicolumn{1}{c}{Language} &       & 1     & 0     & 0     &       & 0     & 2     & 2     & 0     &       & 2     & 1     & 1     &       & 2     & 2     & 2     & 0     &       & 1     & 1     & 1     & \multicolumn{1}{r}{0} \\
\multicolumn{1}{c}{History} &       & 10    & 9     & 12    &       & 23    & 11    & 9     & 11    &       & 11    & 21    & 21    &       & 26    & 25    & 26    & 16    &       & 26    & 25    & 26    & \multicolumn{1}{r}{15} \\
\multicolumn{1}{c}{Philosophy} &       & 16    & 0     & 7     &       & 15    & 6     & 6     & 10    &       & 17    & 15    & 18    &       & 19    & 19    & 18    & 10    &       & 20    & 20    & 22    & \multicolumn{1}{r}{10} \\
\cmidrule{1-1}\cmidrule{3-5}\cmidrule{7-10}\cmidrule{12-14}\cmidrule{16-19}\cmidrule{21-24}\textbf{Total} &       & 169   & 79    & 143   &       & 139   & 105   & 105   & 127   &       & \textbf{296} & \textbf{295} & 288   &       & 246   & 245   & 259   & 174   &       & 279   & 277   & \textbf{291} & 185 \\
\bottomrule
\end{tabular}%
    }
  \label{tab:domains}%
\end{table}%

For each first level category, we gather the top 100 most attended words, as well as those with the highest scores according to each feature selection method. Then, for each word, we look for the corresponding context according to ConceptNet. Table \ref{tab:domains} reports the domain relevance obtained for each category. In BERT and SciBERT, self-attention identifies more domain-relevant words than feature selection methods. However, this is not the case for BioBERT. Recall that in our sample dataset, the set of most attended words produced by BioBERT is the largest (32\%) with respect to the vocabulary, which is a clear indication that the model spreads its attention more widely. Weighing the words by their term frequency (TF), attended words remain more domain-relevant than those obtained through feature selection. In fact, the domain relevance of the frequent attended words is greater or on pair with those selected when TF/IDF is used to weigh the output of feature selection methods: self-attention takes into account not only the importance of words in the document (TF) but also their importance in the document collection (IDF). 

\subsection{Feature evaluation}
To evaluate the quality of the resulting features we measure their stability and their classification performance. Stability is the robustness of a feature subset generated from different training sets from the same distribution \cite{kalousis2007stability}. To measure stability we compute the mean Jaccard coefficient between the different subsets of words generated by each method. We apply 5-fold cross-validation and process each fold with the fine-tuned language models and the feature selection methods. Stability is reported on table \ref{tab:stability}, where we can see that language models attend to the same words with stability values in line with those reported by document count. Attended words are more stable than the rest of the feature selection methods, including chi-square and information gain, which seems to be more volatile across folds. 

\begin{table}[htbp]
  \centering
  \caption{Stability of the features measured using Jackard similarity coefficient}
  \resizebox{0.6\textwidth}{!}{
% Table generated by Excel2LaTeX from sheet 'stability'
\begin{tabular}{|c|c|c|c|c|c|c|}
\hline
 SciBERT & BioBERT & BERT & dc & pd & ig & chi \\
\hline
 0.87  & 0.84  & 0.83  & 0.86  & 0.77  & 0.65  & 0.58 \\
 \hline
\end{tabular}%
}
  \label{tab:stability}
 \end{table}

\begin{figure}[htbp]
     \centering
     \includegraphics[width=\textwidth]{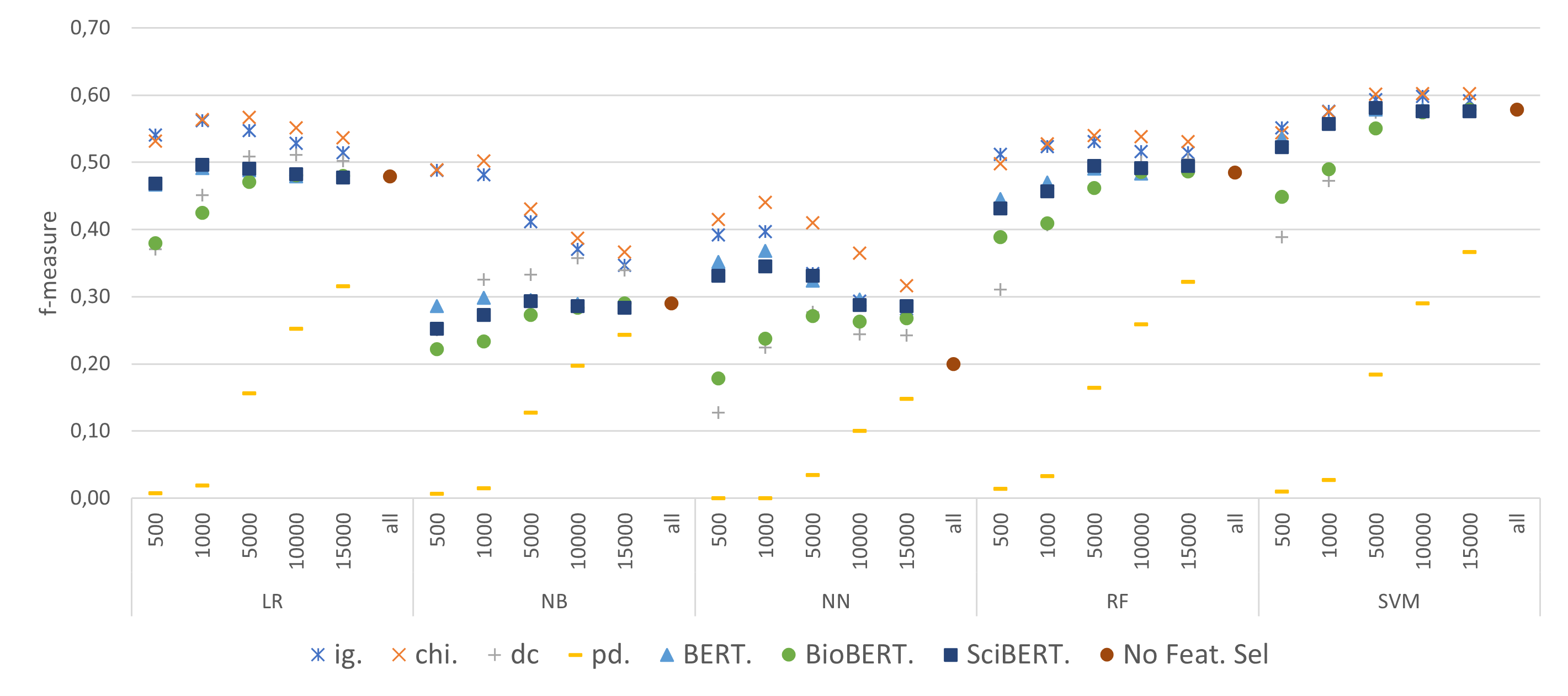}
     \caption{Classifiers performance using distinct feature sets and number of features.}
     \label{fig:my_label}
 \end{figure}

In addition, we use the set of features to learn classifiers for the 22 first level categories using Logistic Regression (LR), Naive Bayes (NB), Random Forest (RF), Neural Networks (NN), and SVM. The neural network comprises an embedding matrix of 100 dimensions and a fully connected layer using sigmoid as activation function. For the SVM the regularization parameter is tuned and for the remaining algorithms we use the recommended settings. We evaluate the classifiers using 5-fold cross validation on the subset of documents where each category was represented with up to 500 papers. The f-measure of the classifiers is shown in figure \ref{fig:my_label}. In general, we observe that traditional feature selection methods like chi-square and information gain mainly help to learn more accurate classifiers than the set of most attended words by the language models. This observation clearly indicates that the success of BERT models in this task is not only driven by the self-attention mechanism but also by the contextualized outputs of the transformer, which are the input of the added classification layer. 
 
\section{Conclusions}

In this paper, we investigate the self-attention mechanism of BERT in a fine-tuning scenario for the classification of scientific articles over a taxonomy of research fields. We observe that attention in the fine-tuned model is focused on words that are highly relevant to the research field of each article. Furthermore, we notice that the most attended words represent just a fraction of the whole vocabulary: a hint that self-attention performs a sort of feature selection. 

We systematically compare the most attended words against those resulting from feature selection methods normally used in text classification. We show that language models and feature selection methods like information gain and chi-square share between 42\% to 55\% of the selected words. We also observe that the attention-based word rankings produced by the transformers are more similar to those obtained using document frequency and information gain. 

From our experiments we conclude that self-attention focuses more on words that are relevant to each research domain than the words produced through conventional feature selection. However, self-attention is not as good to learn classifiers from scratch, especially compared to chi-square and information gain. While self-attention identifies domain-relevant terms the discriminatory information in the fine-tuned model is encoded on the output representations and the additional classification layer. As future work, we plan to investigate the impact of integrating, perhaps as part of the loss function, optimal feature selection methods during fine-tuning of transformer for single sequence classification.

\section*{Acknowledgment}
We gratefully acknowledge the EU Horizon 2020 research and innovation programme under grant agreement No. 825627 (ELG). We also thank Raul Ortega and Cristian Berrio for their contributions to the experimental evaluation.

%The preferred spelling of the word ``acknowledgment'' in America is without an ``e'' after the ``g''. Avoid the stilted expression ``one of us (R. B. G.) thanks $\ldots$''. Instead, try ``R. B. G. thanks$\ldots$''. Put sponsor 

\bibliographystyle{splncs04}
\bibliography{mybibliography}

\end{document}